\documentclass[sigconf]{acmart}
\AtBeginDocument{%
  \providecommand\BibTeX{{%
    \normalfont B\kern-0.5em{\scshape i\kern-0.25em b}\kern-0.8em\TeX}}}

\setcopyright{acmcopyright}
\copyrightyear{2022}
\acmYear{2022}

\acmConference[NICE '23]{}{April 11--14, 2023}{San Antonio, TX, USA}
\begin{document}

\title{Training a spiking neural network on an event-based label-free flow cytometry dataset}


\author{Muhammed Gouda}
\authornote{Both authors contributed equally to this research.}
{\email{MuhammedGoudaAhmed.Gouda@ugent.be}}
\affiliation{%
  \institution{Ghent University - imec}
  \city{Gent}
  \country{Belgium}
}

\author{Steven Abreu}
\authornotemark[1]
\email{s.abreu@rug.nl}
\affiliation{%
  \institution{University of Groningen}
  \city{Groningen}
  \country{Netherlands}
}

\author{Alessio Lugnan}
\email{Alessio.Lugnan@UGent.be}
\affiliation{%
  \institution{Ghent University - imec}
  \city{Gent}
  \country{Belgium}
}

\author{Peter Bienstman}
\email{Peter.Bienstman@UGent.be}
\affiliation{%
  \institution{Ghent University - imec}
  \city{Gent}
  \country{Belgium}
}

\renewcommand{\shortauthors}{Gouda and Abreu, et al.}

\begin{abstract}
Imaging flow cytometry systems aim to analyze a huge number of cells or micro-particles based on their physical characteristics. The vast majority of current systems acquire a large amount of images which are used to train deep artificial neural networks.  However, this approach increases both the latency and power consumption of the final apparatus. In this work-in-progress, we combine an event-based camera with a free-space optical setup to obtain spikes for each particle passing in a microfluidic channel. A spiking neural network is trained on the collected dataset, resulting in 97.7\% mean training accuracy and 93.5\% mean testing accuracy for the fully event-based classification pipeline.

\end{abstract}

\maketitle

\section{Introduction}


Flow cytometry is a technique used to classify different types of cells based on characteristics such as size, shape, or fluorescence. Often, the cells used in flow cytometry are labeled with different biomarkers. Such biomarkers can lead to chemical interactions with the cells that impair the experimental data \cite{doan2018diagnostic}. Label-free, or stain-free, imaging flow cytometry collect images from flowing cells which have not been labelled. These images are used to train a machine learning model for classification. 

The goal of a flow cytometry setup is to classify with high accuracy and low latency (or, equivalently, high throughput). To achieve this, it is necessary to use a sophisticated high-speed camera and a lot of computing power to process large amounts of data. 

Recent work has shown that a neuromorphic event-based camera can achieve low latency with >1,000 frames per second \cite{HeEtAl2022Neuromorphic}. Importantly, the event-based camera does not record full frames at each time step, but rather records only changes in the scene at each pixel. This operating principle inherently filters out temporally redundant information. 

However, in order to use conventional computer vision algorithms on event-based data, it is usually necessary to convert the data into frames. A more natural approach is to process the event-based imaging data using event-based image processing algorithms. We propose the use of spiking neural networks (SNNs) to process the imaging data in a fully event-based and asynchronous way. In this work-in-progress, we show first results that we obtained by training a spiking neural network on event-based flow cytometry data on a GPU. To the best of our knowledge, this are the first results of event-based flow cytometry using spiking neural networks in the literature (Ref. \cite{HeEtAl2022Neuromorphic} uses an artificial neural network, and Ref. \cite{ZhangEtAl2022Work} does not report any performance results). 

\section{Methods}

We proceed by describing the experimental setup in Section \ref{ss:exp-setup} and the spiking neural network we trained in Section \ref{ss:snn}.

\subsection{Imaging flow cytometry}
\label{ss:exp-setup}

\begin{figure}[b]
\centering
\fbox{\includegraphics[width=0.44\textwidth]{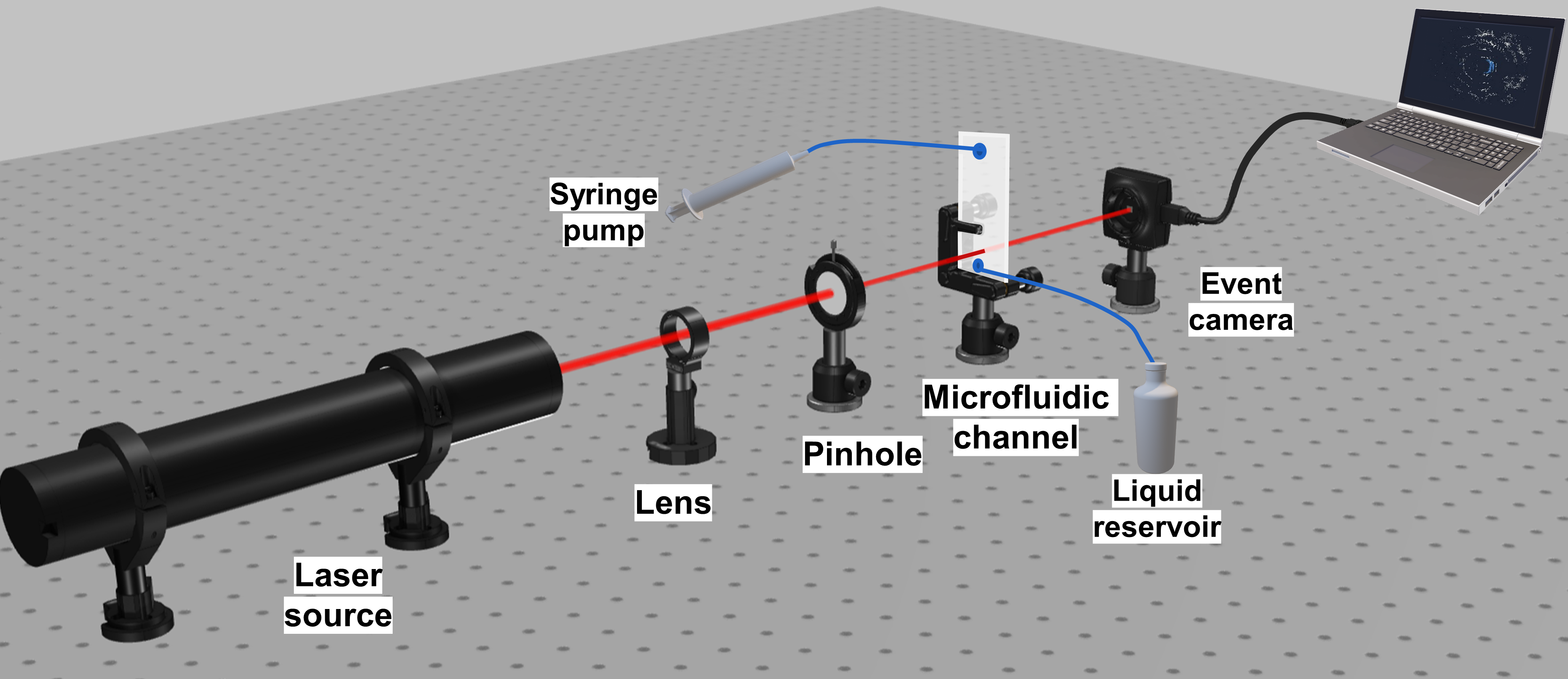}}
\caption{The free-space optical setup built in this work. Light emitted by the 1550 nm He-Ne laser is focused on a PMMA microfluidic channel. Diffraction, scattering and interference of light due to moving particles cause changes in the intensity of light incident on the event sensor.}
\label{fig: sketch3d}
\end{figure}

The experimental setup is shown in Figure \ref{fig: sketch3d}. Here, we send artificial PMMA microbeads of different sizes to a narrow microfluidic channel (of width 200 $\mu m$). The intensity of the light detected by the event-camera stays constant as long as no particle is moving across the field of view. This does not trigger any pixel to fire events. However, once particles enter the field of view,diffraction, scattering and interference of the light trigger firing events which we use later to train our spiking neural network. The fact that we obtain events only in the presence of particles significantly improves the memory efficiency of the overall system. The size of the dataset we get with our system is on the order of tens of gigabytes, whereas in a previous work using a traditional frame-based camera, the dataset was on the order of hundreds of gigabytes \cite{lugnan2020machine}. 


We used two different classes of spherical microparticles (class A of diameter 16 µm and class B with a diameter of 20 µm). We ran four separate experiments for each class of microparticles, where each experiment ran for $T_{exp} \approx 60 s$ for a total of 480 seconds of data. The accumulation time for a single particle is $T_{acc} \approx 10 ms$. Therefore, we have around 6,000 samples per experiment, or 24,000 samples in total per class. 
We train the network for four different train-test splits, each split using a different experiment as testing data, and the remaining experiments as training data.

\subsection{Spiking neural network}
\label{ss:snn}

We pre-process our event-based imaging data using the Tonic library \cite{tonic2021}. The pre-processing involved a spatial downsampling from $640 \times 480 \times 2$ to $32 \times 24 \times 2$ (event polarity is left unchanged), and a temporal downsampling by passing the events for each neuron corresponding to a downsampled pixel of fixed polarity through a discretized leaky-integrate-and-fire (LIF) neuron:
\begin{align*}
    s^{out}_i(n+1) &= 1 \text{ if } \left( \beta u_i(n) + w s^{in}_i(n+1) \right) \geq u_{thr} \text{ else } 0 \\
    r_i(n+1) &= \max_{t \in \{ 0, \ldots, t_{rf} \} } s^{out}_i(n+1-t) \\
    u_i(n+1) &= 0 \text{ if } \left( r_i(n+1) \right) \text{ else } \left( \beta u_i(n) + w s^{in}_i(n+1) \right)
\end{align*}
where $u_i(n)$, $s^{out}_i(n)$, $r_i(n)$ denote the membrane potential, binary spike output, and binary refractory period of neuron $i$ at timestep $n$, respectively. 
Further, $\beta = 0.9$ is the membrane decay rate, $w = 1.0$ is the synaptic weight, $u_{thr}=3.0$ is the threshold voltage, $t_{rf} = 2$ is the refractory period, $s^{in}_i(n)$ equals unity if there is an input event from the event-based camera in the $i$th $20 \times 20 \times 1$ patch of pixels at timestep $n$ (and zero otherwise).

For classification, we use a feedforward network of LIF neurons with an input layer of size $32 \times 24 \times 2=1536$, a hidden layer of size $100$, and an output layer of size $20$ (using population coding with $10$ neurons per output class). The neuron parameters are constant for all neurons in the network, with the membrane decay rate $\beta=0.9$ and membrane threshold $U_{thr}=0.5$. 
We use the Adam optimizer to optimize the mean squared error on the output spike rate, with a desired population spike rate of $20\%$ (correct) and $80\%$ (incorrect). Backpropagation is applied through a shifted Heaviside function in the forward pass and a fast sigmoid 
\begin{align*}
    S = \frac{U}{1+k\|U\|}
\end{align*}
with slope $k=75$ in the backward pass \cite{ZenkeGanguli2018}. We use snnTorch \cite{EshraghianEtAl2021Training} for our implementation. 

\section{Results}

\begin{figure}[H]
\centering
\includegraphics[width=0.47\textwidth]{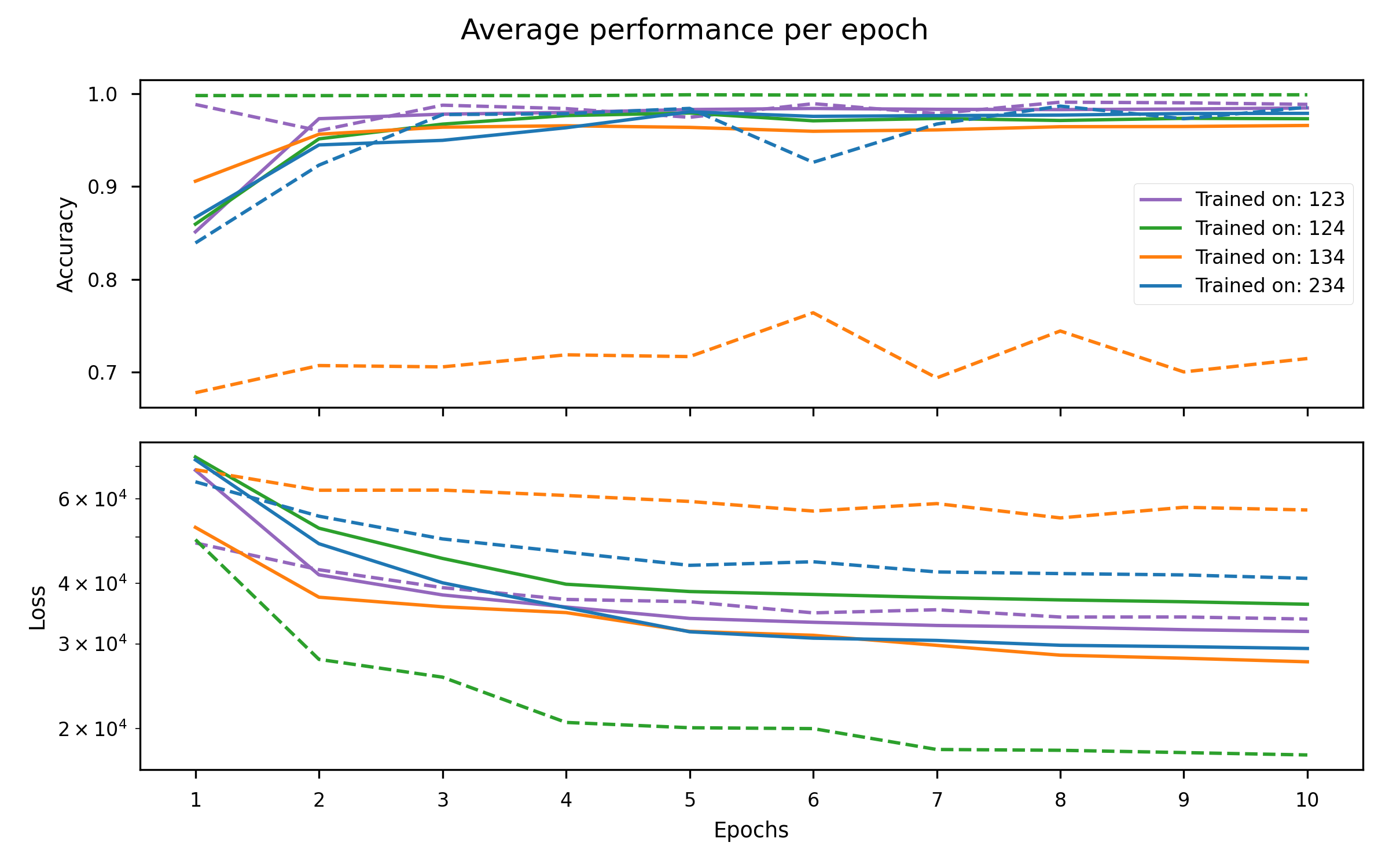}
\caption{Performance of the spiking neural network trained with backpropagation for ten epochs. Dashed lines show testing, solid lines show training. Different lines indicate which experiments were used for training, the remaining experiment was used for testing only.}
\label{fig:snn-performance}
\end{figure}

We trained the spiking neural network described in Section \ref{ss:snn} for 10 epochs on a NVIDIA GeForce RTX 2080 Ti GPU with 11 GB of memory. Figure \ref{fig:snn-performance} show the training and testing performance for this experiment. The figure shows that the accuracy exceeds 98.5\% for all experiments, with the exception of the testing accuracy for the network trained on experiments 1, 3, and 4. More work is needed to analyze the anomalous performance on this data split.

\section{Outlook}
\label{s:outlook}

As next steps, we are investigating the use of exact timing information for a time-to-first-spike classification that should yield even faster classification. In this work, we fixed the hyperparameters use to the values mentioned in Section \ref{ss:snn}. For further performance gains, we will optimize the hyperparameters and network architecture through a cross-validation procedure. Finally, we will transfer the SNN to a neuromorphic chip for a fully neuromorphic processing pipeline.

\begin{acks}
This work was performed in the context of the European projects Neoteric (grant agreement 871330), Prometheus (grant agreement 101070195) and Post-Digital project (grant agreement 860360). 
\end{acks}

\bibliographystyle{ACM-Reference-Format}
\bibliography{sample-base}


\end{document}